\documentclass[letterpaper]{article} 
\usepackage{aaai2027}  
\usepackage{helvet}  
\usepackage{courier}  
\usepackage[hyphens]{url}  
\usepackage{graphicx} 
\usepackage{natbib}  
\usepackage{caption} 
\usepackage{algorithm}
\usepackage{algorithmic}

\usepackage{newfloat}
\usepackage{listings}
\DeclareCaptionStyle{ruled}{labelfont=normalfont,labelsep=colon,strut=off} 
\floatstyle{ruled}
\newfloat{listing}{tb}{lst}{}
\floatname{listing}{Listing}

\usepackage{booktabs}

\usepackage[utf8]{inputenc} 
\usepackage[T1]{fontenc}    
\usepackage{url}            
\usepackage{booktabs}       
\usepackage{amsfonts}       
\usepackage{nicefrac}       
\usepackage{microtype}      
\usepackage{xcolor}         
\usepackage{graphicx}
\usepackage{booktabs}
\usepackage{multirow}
\usepackage{amsmath} 

\usepackage{placeins}

\usepackage{subcaption}
\usepackage{caption}

\nocopyright

\title{MiMaAlign: Hierarchical Micro-Macro Alignment for Physically Plausible Video Generation}
\author{
    Yansong Liu\textsuperscript{\rm 1,2}\thanks{Work done when Yansong Liu was an intern at Taobao \& Tmall Group of Alibaba.},
    Litong Gong\textsuperscript{\rm 2},
    Weijie Li\textsuperscript{\rm 2},
    Tiezheng Ge\textsuperscript{\rm 2},
    Ming Zeng\textsuperscript{\rm 1}
}
\affiliations{
    \textsuperscript{\rm 1} Xiamen University, Xiamen, China\\
    \textsuperscript{\rm 2}  Alibaba Group, Beijing, China

    \{liuyansong@stu., zengming@\}xmu.edu.cn, \{gonglitong.glt, weijie.lwj0,  tiezheng.gtz\}@alibaba-inc.com
}

\begin{document}

\maketitle

\begin{abstract}
Text-to-video (T2V) diffusion models have made remarkable progress in generating high-fidelity visuals, yet their outputs still often exhibit physically implausible phenomena, including discontinuous dynamics and implausible event interactions. Existing representation alignment methods mainly transfer generic video features or short-range motion cues, treating physical plausibility as a single-level representation problem. To address this issue, we propose MiMaAlign, a hierarchical micro-macro alignment framework that transfers complementary dynamics-oriented and event-oriented knowledge into different abstraction depths of a pretrained video diffusion model. Specifically, micro-level continuous dynamics alignment imposes constraints by aligning diffusion features with a continuous-time motion representation learned by a Neural Ordinary Differential Equation (Neural ODE), while macro-level event-commonsense alignment provides representations of object interactions and event semantics through an instruction-prompted vision-language model (VLM). With training-only auxiliary objectives, we align the complementary supervision from the two teachers with shallower and deeper generator features, respectively, enabling the generator to internalize the corresponding physical priors. Empirical evaluations show that, on CogVideoX-2B, MiMaAlign achieves a relative improvement of up to 21.0\% in the Physical Commonsense (PC) score on the VideoPhy benchmark over the baseline, and also delivers substantial gains on the highly challenging VideoPhy2 benchmark, while maintaining general video quality on VBench and being supported by blind user comparisons.
\end{abstract}

\section{Introduction}
\label{sec:intro}

\begin{figure*}[ht]
    \centering
    \includegraphics[width=0.88\textwidth]{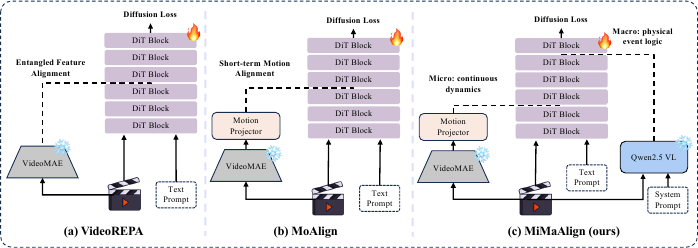} 
    \vspace{-1em}
    \caption{A comparison of alignment frameworks. (a) VideoREPA uses a single teacher with entangled features. (b) MoAlign disentangles short-term motion for alignment. (c) Our MiMaAlign employs hierarchical micro-macro alignment to synergistically enhance physical plausibility.}
    \label{fig:comparison}
    \vspace{-1em}
\end{figure*}

In recent years, Video Diffusion Models (VDMs) have made substantial progress, synthesizing dynamic content with high visual fidelity and showing strong potential in applications such as content creation, virtual reality, and digital storytelling. Representative systems like LongCat-Video~\cite{LongCat25}, Wan2.1~\cite{Wan25}, CogVideoX~\cite{Yang25b}, HunyuanVideo~\cite{Lab25}, Cosmos~\cite{NVIDIA25}, and Open-Sora Plan~\cite{Lin24a}, which are based on Diffusion Transformers (DiTs) and trained on large-scale video-text data, have achieved impressive levels of visual quality and model scalability. However, despite near-photorealistic single-frame appearance and local details~\cite{Huang24, Zheng25}, recent physics benchmarks show that current mainstream Text-to-Video (T2V) models still exhibit a clear gap in simulating the physical laws of the real world~\cite{Bansal25a, Bansal25b, yuan2025likephys, kang2024far}. A key reason is that physical plausibility in video is not a single-level problem, but manifests simultaneously at two complementary levels. At the microscopic level, object motion trajectories in generated videos are often discontinuous and non-smooth, frequently violating gravity or the conservation of momentum, and lacking the intrinsic continuity of physical dynamics~\cite{Bear2021Physion}. At the macroscopic level, models struggle to accurately understand and reproduce complex physical events, exhibiting erroneous interaction logic between objects, confused causal relationships, or events that as a whole defy physical common sense~\cite{schmidt2025video, komanduri2025causalvlbench}. These deficiencies in microscopic dynamics and macroscopic event logic jointly undermine the realism and credibility of the generated content, severely limiting its value in advanced applications that demand high reliability.

Recent works have begun to explore transferring knowledge from external encoders through representation alignment (REPA~\cite{Yu25}), offering a lightweight way to improve pretrained generators. VideoREPA~\cite{Zhang25b} applies this idea to video generation through token relation distillation from video foundation models, while MoAlign~\cite{Bhowmik25} further separates motion-oriented features using optical flow supervision. These methods demonstrate the value of representation transfer, but they still address physical realism mainly through a single teacher space. Generic video features may retain substantial appearance information, whereas flow-supervised features emphasize local displacement. Neither representation explicitly complements motion evolution with event-level information about object interactions, causal dependencies, and plausible outcomes.


We therefore formulate physical plausibility in video generation as a hierarchical micro-macro representation problem. The micro level concerns how the visual state evolves over time, for which the teacher should be aware of multi-step dynamics rather than merely adjacent-frame displacements. The macro level concerns whether an interaction forms a coherent physical event, for which the teacher should capture objects, relations, and event semantics. Prior studies have shown that representations at different layers of deep visual models exhibit different abstraction preferences: shallower layers tend to capture local details, while deeper layers tend to encode coarse-grained global structures~\cite{reed25, szegedy2016rethinking, sohl2015deep}. Motivated by this, we propose our central hypothesis: these two kinds of knowledge are complementary and should supervise different abstraction depths of the generator---dynamics-oriented representations correspond to local, fine-grained motion evolution and should act on shallower layers, while event-oriented representations correspond to global, coarse-grained event structures and should act on deeper layers. Based on this hypothesis, we propose MiMaAlign, a hierarchical micro-macro representation alignment framework for physically plausible video generation. At the micro level, we leverage a Neural ODE~\cite{Chen2018NeuralODE} to learn continuous-time motion representations, capturing dynamic priors that better conform to real physical evolution. Unlike discrete short-term motion modeling based on optical flow, this approach is better suited to describing the frame-to-frame continuous, fine-grained motion evolution governed by dynamical laws~\cite{park2020vid}. Our evaluation on Physion (as shown in Table~\ref{tab:physion_motion}) shows that the motion representations learned via the Neural ODE significantly outperform baselines supervised by optical flow on the OCP task. At the macro level, we capture object, interaction, and event semantics from an instruction-prompted Vision-Language Model (VLM)~\cite{schmidt2025video}. By hierarchically integrating these two complementary sources of physical knowledge within a unified framework, MiMaAlign synergistically improves the realism of generated videos in both micro-level dynamics and macro-level event logic, producing videos that better conform to the physical world. Figure~\ref{fig:comparison} contrasts existing single-teacher alignment paradigms with our approach.

\begin{table}[t]
\small
\centering
\setlength{\tabcolsep}{2.0pt} 
\begin{tabular}{lcccc}
\toprule
Methods & link & roll & contain & dominoes \\
\midrule
VideoMAEv2 (Optical Flow) &
59.9\% & 51.6\% & 60.8\% & 69.2\% \\
VideoMAEv2 (Neural ODE) &
\textbf{65.0\%} & \textbf{59.2\%} & \textbf{64.1\%} & \textbf{71.7\%} \\
\bottomrule
\end{tabular}
\vspace{-1em}
\caption{Physics understanding evaluation on Physion~\cite{Bear2021Physion}. Our proposed Neural-ODE-based continuous dynamics modeling significantly outperforms optical-flow-based short-term motion modeling on the Object Contact Prediction (OCP) task.}
\vspace{-2em}
\label{tab:physion_motion}
\end{table}

Our contributions are summarized as follows: 
\begin{itemize}
    \item We formulate physically plausible video generation as a two-level micro-macro representation problem that distinguishes continuous dynamics from event-level coherence, offering a more targeted alternative perspective to previous single-teacher alignment.
    \item We propose MiMaAlign, a hierarchical dual-teacher representation alignment framework that learns micro-level motion representations via Neural-ODE-based continuous-time feature prediction and extracts macro-level event representations from an instruction-prompted vision-language model, aligning their complementary supervision with shallower and deeper layers of the generator through training-only auxiliary objectives.
    \item We evaluate this design on VideoPhy, VideoPhy2, VBench, and blind human comparisons. MiMaAlign shows clear advantages over comparable fine-tuning and representation-alignment baselines, and the gains transfer across the evaluated T2V backbones.
\end{itemize}

\section{Related Work}
\textbf{Video Diffusion Models}
Video Diffusion Models (VDMs) have advanced quickly and now achieve strong results in video synthesis. Early approaches extended U-Net with temporal attention to handle video sequences~\cite{Ho22, Blattmann23b}, but they often had difficulty maintaining long-range temporal coherence. More recent work has therefore moved to Transformer-based backbones. Current models, including LongCat-Video~\cite{LongCat25}, CogVideoX~\cite{Yang25b}, Wan2.1~\cite{Wan25}, HunyuanVideo~\cite{Lab25}, and Open-Sora Plan~\cite{Lin24a}, are largely built on the Diffusion Transformer (DiT) architecture and trained on large-scale video-text datasets, leading to substantial gains in visual fidelity and generation diversity. Beyond content creation, recent systems such as Cosmos~\cite{NVIDIA25} further position VDMs as world simulators for Physical AI. However, despite the near-photorealistic quality of individual frames, these models still frequently produce dynamics that violate physical laws, such as discontinuous motion trajectories, failed object interactions, and outcomes that defy basic physical commonsense, which limits their reliability in applications that demand faithful physical behavior.

\textbf{Physically Plausible Video Generation}
Recent benchmarks show that state-of-the-art text-to-video models still struggle to respect real-world physics~\cite{Bansal25a, Bansal25b, Huang24}. Prior studies either inject explicit physical priors through simulation or 3D/4D representations~\cite{Liu24, Xie24, Zhang24, Montanaro24, Lin24b, Lin24c, Xie25}, which rely on explicit 3D assets or scene-specific simulators, or guide generation via LLM-based prompt refinement and physics-oriented rewards~\cite{Xue25, Wang25, Lin25}, which act only on the text or reward level without directly constraining the generator's internal spatiotemporal dynamics. A more lightweight and model-agnostic route is representation alignment (REPA)~\cite{Yu25, Leng25, Tian2026UREPA, Wu25REG, wang2025repa, Wang25Disperse, Yang26DiverseDiT, Chefer26SelfFlow, reed25}, which matches intermediate diffusion features to representations from a pre-trained encoder to distill useful knowledge. In video generation, VideoREPA~\cite{Zhang25b} transfers spatiotemporal and physical cues from video foundation models~\cite{Wang23, bardes2023vjepa, girdhar2023omnimae, tong2022videomae} through token-relation distillation, and MoAlign~\cite{Bhowmik25} learns a disentangled motion subspace via optical flow prediction. However, these methods all rely on single-level cues from a single teacher space, such as entangled appearance features or short-term pixel displacements, and fail to jointly model continuous-time dynamical evolution and macro-level event-level physical commonsense. MiMaAlign builds on this line of work; its distinction lies in decomposing physical plausibility into two complementary failure modes, micro-level dynamics and macro-level event logic, constructing a dedicated teacher for each---a Neural-ODE-based continuous-time dynamics teacher and an instruction-prompted VLM event-semantics teacher---and aligning the two complementary supervisions with the shallower and deeper layers of the video generator through a hierarchical alignment strategy.

\begin{figure*}[ht]
    \centering
    \includegraphics[width=0.88\textwidth]{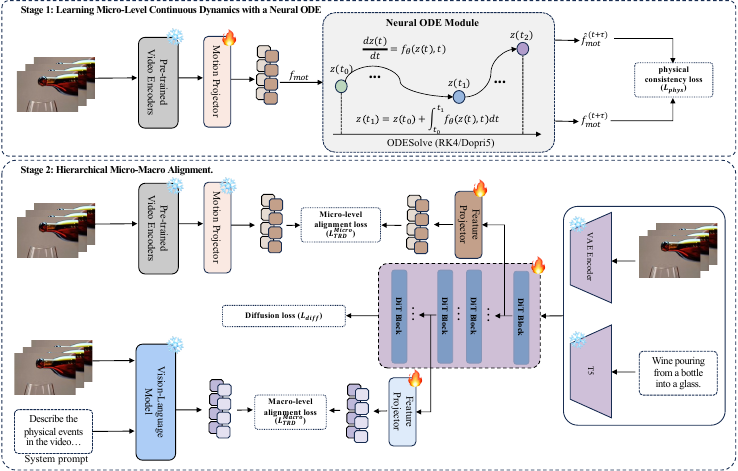} 
    \vspace{-1em}
    \caption{Overview of the proposed MiMaAlign framework. \textbf{Stage 1}: A Neural-ODE-based micro-dynamics teacher is trained to model continuous-time motion dynamics. \textbf{Stage 2}: The complementary supervision from the micro-dynamics teacher and the VLM-based macro-event teacher is aligned with shallower and deeper layers, respectively, to improve physical plausibility.}
    \label{fig:pipeline}
    \vspace{-1em}
\end{figure*}

\section{Method}

\subsection{Preliminaries}
\textbf{Video diffusion models.} Video Diffusion Models (VDMs) are commonly trained in the latent space of a pre-trained video autoencoder (e.g., a 3D-VAE), which reduces the cost of diffusion in pixel space and often improves training stability. Given a video clip $x_0 \in \mathbb{R}^{F \times H \times W \times 3}$ with $F$ frames, the encoder maps it to a latent tensor $z_0 = \mathcal{E}(x_0) \in \mathbb{R}^{f \times h \times w \times C}$. The forward diffusion process then adds Gaussian noise to this latent in a sequence of steps. At timestep $t$, the noisy latent is
\begin{equation}
z_t = \alpha_t z_0 + \sigma_t \epsilon,\quad \epsilon \sim \mathcal{N}(0, I),
\end{equation}
where $(\alpha_t,\sigma_t)$ are set by a noise schedule. Training fits a conditional denoiser $\epsilon_\theta(\cdot)$, typically implemented as a Diffusion Transformer (DiT), to predict the injected noise $\epsilon$ from $z_t$, $t$, and the text condition $c$. The standard objective is the mean squared error (MSE):
\begin{equation}
\mathcal{L}_{\text{diff}}=\mathbb{E}_{z_0,t,\epsilon}\big[\|\epsilon-\epsilon_\theta(z_t,t,c)\|_2^2\big].
\end{equation}
At inference time, the model starts from Gaussian noise $z_T \sim \mathcal{N}(0,I)$ and iteratively denoises to recover $z_0$, which is decoded into a video $\hat{x}_0 = \mathcal{D}(z_0)$ by the decoder $\mathcal{D}$. Because video has higher spatiotemporal dimensionality than images, VDMs often add spatiotemporal attention, factorized spatiotemporal modules, or multi-scale designs to better model inter-frame dependencies and long-range consistency while preserving visual quality.

\textbf{Representation Alignment (REPA).}
Representation alignment (REPA) is a regularization technique that aligns the hidden states of a diffusion model with features from a pre-trained visual encoder, such as DINOv2~\cite{Oquab2023DINOv2}. This process can accelerate convergence and improve the quality and controllability of the generated output.

REPA~\cite{Yu25} was first proposed for image diffusion models. Given an input image $x$, an encoder $f(\cdot)$ produces token features $Y^{\ast}=f(x)\in\mathbb{R}^{N\times D}$. The diffusion model's intermediate hidden representation $h_t$ is mapped by a lightweight projection head $P_\phi(\cdot)$ into the teacher's feature space, yielding $\tilde{H}=P_\phi(h_t)$. The alignment loss then maximizes the token-wise cosine similarity:
\begin{equation}
\mathcal{L}_{\text{REPA}}
= -\mathbb{E}\left[\frac{1}{N}\sum_{n=1}^{N}\mathrm{sim}(Y_n^{\ast},\tilde{H}_n)\right],
\end{equation}
where $\mathrm{sim}(\cdot,\cdot)$ denotes the cosine similarity. The total loss combines this term with the standard diffusion objective:
\begin{equation}
\mathcal{L}_{\text{total}}=\mathcal{L}_{\text{diff}}+\lambda\,\mathcal{L}_{\text{REPA}},
\end{equation}
where $\lambda$ is a weighting hyperparameter. While the original REPA focuses on spatial semantics in images, later works like VideoREPA~\cite{Zhang25b} and MoAlign~\cite{Bhowmik25} have extended it to video by aligning spatiotemporal token relations or features from a dedicated motion subspace. This improves temporal consistency and physical plausibility without relying on an explicit physics engine.

\textbf{Neural Ordinary Differential Equations.}
Conventional sequence models, such as recurrent neural networks, process data through discrete state transitions and thus struggle to accurately model dynamical systems that evolve in continuous time. Neural ODEs~\cite{Chen2018NeuralODE} offer an alternative by framing the evolution of a system's hidden state as a continuous transformation. They learn an ordinary differential equation parameterized by a neural network to describe the system's dynamics.

A Neural ODE defines the time derivative of a hidden state $z(t)$ using a neural network $f_\theta$:
\begin{equation}
\frac{dz(t)}{dt}=f_\theta(z(t),t).
\end{equation}
Given an initial state $z(t_0)$, the state at a future time $t_1$ is found by integrating this equation:
\begin{equation}
z(t_1)=z(t_0)+\int_{t_0}^{t_1} f_\theta(z(t),t)\,dt.
\end{equation}
In practice, this integral is approximated using a black-box numerical ODE solver, denoted as $\text{ODESolve}(\cdot)$. Since physical dynamics in videos are continuous processes observed at discrete time steps (frames), Neural ODEs are well-suited for modeling and disentangling continuous, fine-grained motion dynamics from video data.


\subsection{Hierarchical Micro-Macro Alignment Framework}
To improve the physical plausibility of video diffusion models (VDMs), we propose MiMaAlign, a two-stage hierarchical physical alignment framework, as illustrated in Figure~\ref{fig:pipeline}. First, we train a micro-level motion teacher with a Neural ODE to capture continuous dynamics. Second, we adopt a vision-language model (VLM) as a macro-level event teacher to distill high-level physical reasoning and interaction commonsense. Finally, a hierarchical alignment strategy is employed to distill their complementary knowledge into the VDM.


\begin{table*}[t]
\small
\centering
\setlength{\tabcolsep}{7pt}
\renewcommand{\arraystretch}{1}
\begin{tabular*}{0.88\textwidth}{@{\extracolsep{\fill}}lcccccccccc}
\toprule
\multirow{3}{*}{Methods} &
\multicolumn{8}{c}{VideoPhy} &
\multicolumn{2}{c}{VideoPhy2} \\
\cmidrule(lr){2-9}\cmidrule(lr){10-11}
&
\multicolumn{2}{c}{Solid--Solid} &
\multicolumn{2}{c}{Solid--Fluid} &
\multicolumn{2}{c}{Fluid--Fluid} &
\multicolumn{2}{c}{Overall} &
\multirow{2}{*}{SA} & \multirow{2}{*}{PC} \\
\cmidrule(lr){2-3}\cmidrule(lr){4-5}\cmidrule(lr){6-7}\cmidrule(lr){8-9}
&
SA & PC & SA & PC & SA & PC & SA & PC & & \\
\midrule
HunyuanVideo &
\textbf{55.2} & \underline{16.1} &
70.5 & \underline{32.2} &
63.6 & \textbf{54.5} &
63.1 & \underline{29.1} &
26.8 & 66.8 \\
LongCat-Video &
50.4 & \underline{16.1} &
73.3 & 30.1 &
\underline{69.1} & 40.0 &
63.1 & 25.9 &
26.6 & 67.0 \\
Cosmos-Predict2.5-2B &
32.2 & 4.2 &
54.1 & 15.8 &
41.8 & 23.6 &
43.0 & 12.2 &
21.4 & 56.1 \\
Wan2.1-14B &
49.7 & 15.4 &
71.2 & 27.4 &
\textbf{72.7} & 43.6 &
62.5 & 25.0 &
\underline{27.5} & 59.2 \\
\midrule
CogVideoX-2B &
46.9 & 13.3 &
70.5 & 30.1 &
65.5 & 49.1 &
59.9 & 26.2 &
27.0 & 64.2 \\
+ Finetune &
49.7 & 15.4 &
71.9 & 29.5 &
65.5 & 47.3 &
61.6 & 26.5 &
27.0 & 62.7 \\
+ VideoREPA &
51.0 & \underline{16.1} &
\underline{75.3} & 31.5 &
67.3 & 47.3 &
64.0 & 27.6 &
27.0 & 64.8 \\
+ MoAlign &
51.0 & 13.3 &
\underline{75.3} & 30.8 &
\underline{69.1} & \underline{50.9} &
\underline{64.2} & 26.7 &
26.3 & \underline{67.5} \\
+ MiMaAlign (ours) &
\underline{53.1} & \textbf{18.9} &
\textbf{76.7} & \textbf{35.6} &
63.6 & \textbf{54.5} &
\textbf{64.8} & \textbf{31.7} &
\textbf{27.6} & \textbf{70.0} \\
\bottomrule
\end{tabular*}
\vspace{-1em}
\caption{Quantitative comparison on VideoPhy and VideoPhy2 benchmarks. SA denotes Semantic Adherence and PC denotes Physical Commonsense. Best results are in \textbf{bold}, and second best are \underline{underlined}.}
\vspace{-2em}
\label{tab:main_results}
\end{table*}

\paragraph{Stage 1: Learning Micro-Level Continuous Dynamics with a Neural ODE.}
The goal of this stage is to learn a feature representation that encodes continuous-time, fine-grained motion dynamics. This is challenging because features from pre-trained video encoders typically entangle motion with static appearance. Furthermore, discrete signals such as optical flow used in prior work~\cite{Bhowmik25} cannot capture the smooth, continuous trajectories characteristic of real-world physical motion.

To address this, we adopt a self-supervised framework based on a Neural ODE to distill continuous motion dynamics from these entangled features. Given a video clip $x_0\in\mathbb{R}^{F \times H \times W \times 3}$, a frozen video encoder $\mathcal{V}$ (e.g., VideoMAEv2~\cite{Wang23}) extracts spatiotemporal features $S=\mathcal{V}(x_0)\in\mathbb{R}^{F' \times H' \times W' \times D_v}$. A learnable projection head $\Phi_\psi$ maps these into a latent motion space:
\begin{equation}
f_{\text{micro}}=\Phi_\psi(S)\in\mathbb{R}^{F' \times H' \times W' \times D_m},\quad D_m \ll D_v.
\end{equation}
We model the evolution of $f_{\text{micro}}$ with a Neural ODE, where a network $\mathcal{F}_\theta$ parameterizes the dynamics. Using a numerical integrator, we predict the state at a future offset $\tau$ from time $t$:
\begin{equation}
\hat{f}_{\text{micro}}^{(t+\tau)}=\text{ODESolve}\big(\mathcal{F}_\theta,f_{\text{micro}}^{(t)},\tau\big),\quad \tau=1,\dots,N.
\end{equation}
The model is trained with a self-supervised objective that minimizes the error between predicted and observed features:
\begin{equation}
\mathcal{L}_{\text{phys}}=
\sum_{t=N}^{F'-N}\sum_{\tau=1}^{N}
\big\| f_{\text{micro}}^{(t+\tau)}-\hat{f}_{\text{micro}}^{(t+\tau)} \big\|_2^2.
\end{equation}
Minimizing $\mathcal{L}_{\text{phys}}$ forces the model to learn the underlying continuous dynamics~\cite{Zhan25}, yielding a micro-level motion feature sequence $F_{\text{Micro}}=f_{\text{micro}}$.

\textbf{Stage 2: Hierarchical Micro-Macro Alignment.}
While the micro-level motion teacher constrains continuous dynamics, it cannot characterize object interactions or event-level physical logic. Motivated by this, we further introduce a macro-level physical event teacher built upon a VLM, and jointly inject the complementary micro- and macro-level physical knowledge into the VDM via token relation distillation.

We use a large, frozen vision-language model $\mathcal{V}_{\text{VLM}}$ (e.g., Qwen2.5-VL~\cite{bai2025qwen}) as the macro-level physical event teacher. Given a video clip $x_0$ and a text prompt $I$, we obtain the final-layer hidden states of the VLM and select only the tokens corresponding to the visual input patches to form the macro-level target $F_{\text{Macro}} = \text{Vis}(\mathcal{V}_{\text{VLM}}(x_0, I)) \in \mathbb{R}^{F'' \times H'' \times W'' \times D_s}$, where $\text{Vis}(\cdot)$ denotes selecting and reshaping the visual tokens. Next, we take the hidden representations $H^{(i)}$ and $H^{(j)}$ from a shallow layer $i$ and a deep layer $j$ of the VDM, and employ two lightweight projection networks $P_i$ and $P_j$ to map them into spaces compatible with the teacher feature dimensions: $Z_{\text{Micro}} = P_i(H^{(i)}) \in \mathbb{R}^{F' \times H' \times W' \times D_m}$ and $Z_{\text{Macro}} = P_j(H^{(j)}) \in \mathbb{R}^{F'' \times H'' \times W'' \times D_s}$.

Subsequently, we employ token relation distillation (TRD) to transfer the relational structure information within the features. For any feature tensor $F$, we define a unified token relation matrix over spatiotemporal token indices $p=(d,u)$ and $q=(e,v)$ as $S^{p,q}(F)=\mathrm{sim}\big(F^{(p)},F^{(q)}\big)$, where $\mathrm{sim}(\cdot,\cdot)$ denotes the cosine similarity: when $d=e$, the entry captures the spatial relation between tokens $u$ and $v$ within frame $d$; when $d\neq e$, it captures the temporal relation between tokens across frames. The TRD loss achieves alignment by minimizing the L1 distance between the relation matrices of the student features $Z$ and the teacher features $F$:
\begin{equation}
\mathcal{L}_{\text{TRD}}(Z, F)=\sum_{p,q}\big\|S^{p,q}(Z)-S^{p,q}(F)\big\|_1 .
\end{equation}
Building on this, we define the micro-level alignment loss $\mathcal{L}_{\text{TRD}}^{\text{Micro}} = \mathcal{L}_{\text{TRD}}(Z_{\text{Micro}}, F_{\text{Micro}})$ and the macro-level alignment loss $\mathcal{L}_{\text{TRD}}^{\text{Macro}} = \mathcal{L}_{\text{TRD}}(Z_{\text{Macro}}, F_{\text{Macro}})$, and impose them at different levels of the VDM, forming a hierarchical supervision strategy that promotes the injection of complementary knowledge. 

Finally, our overall training objective is formulated as a weighted sum of the standard diffusion loss and the two TRD alignment losses:
\begin{equation}
\mathcal{L}_{\text{MiMaAlign}} = \mathcal{L}_{\text{diff}} + \lambda_{\text{Micro}}\mathcal{L}_{\text{TRD}}^{\text{Micro}} + \lambda_{\text{Macro}}\mathcal{L}_{\text{TRD}}^{\text{Macro}},
\end{equation}
where $\lambda_{\text{Micro}}$ and $\lambda_{\text{Macro}}$ are hyperparameters controlling the intensity of the respective knowledge alignment.

\begin{figure*}[t]
    \centering
    \vspace{-1em}
    \includegraphics[width=0.88\textwidth]{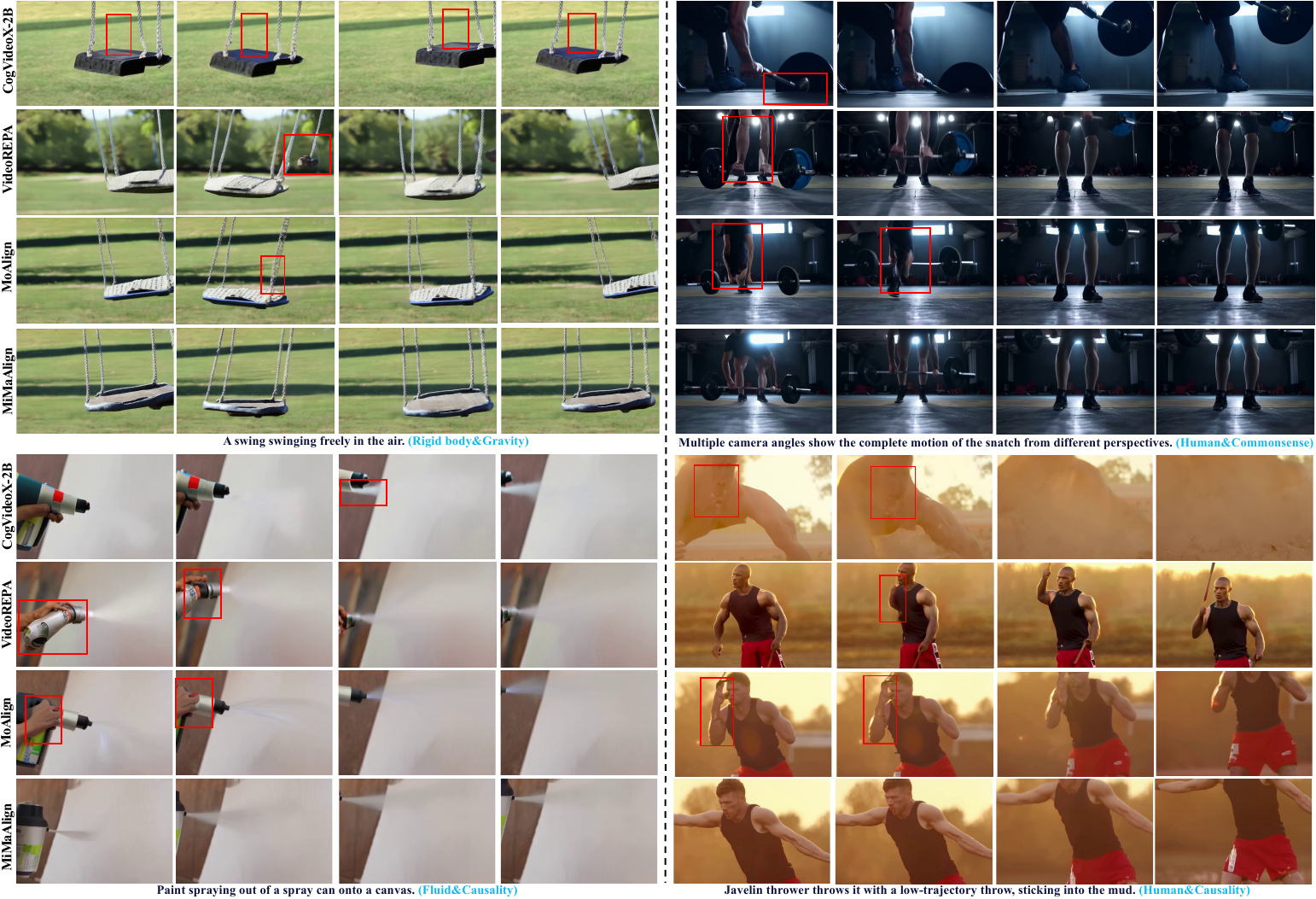} 
    \vspace{-1em}
    \caption{Qualitative comparison of CogVideoX, VideoREPA, MoAlign, and MiMaAlign (Ours).}
    \vspace{-1em}
    \label{fig:qualitative}
\end{figure*}

\begin{table*}[!t]
\small
\centering
\setlength{\tabcolsep}{2pt}
\renewcommand{\arraystretch}{1}
\begin{tabular*}{0.88\textwidth}{@{\extracolsep{\fill}}lcccccccc}
\toprule
\multirow{2}{*}{Methods} & 
\multicolumn{1}{c}{Aesthetic} & 
\multicolumn{1}{c}{Background} & 
\multicolumn{1}{c}{Imaging} & 
\multicolumn{1}{c}{Motion} & 
\multicolumn{1}{c}{Dynamic} & 
\multicolumn{1}{c}{Subject} & 
\multicolumn{1}{c}{Temporal} & 
\multicolumn{1}{c}{Quality} \\
& 
\multicolumn{1}{c}{Quality} & 
\multicolumn{1}{c}{Consistency} & 
\multicolumn{1}{c}{Quality} & 
\multicolumn{1}{c}{Smoothness} & 
\multicolumn{1}{c}{Degree} & 
\multicolumn{1}{c}{Consistency} & 
\multicolumn{1}{c}{Flickering} & 
\multicolumn{1}{c}{Score} \\
\midrule
CogVideoX-2B & \underline{60.7} & 95.0 & \underline{60.5} & 97.9 & \textbf{63.9} & 92.0 & 97.6 & 80.6 \\
+VideoREPA & 59.7 & 95.4 & 59.3 & \textbf{98.5} & \textbf{63.9} & \underline{93.0} & \textbf{98.2} & \underline{81.1} \\
+MoAlign   & \underline{60.7} & \underline{95.5} & 59.6 & \underline{98.4} & 59.7 & \textbf{93.6} & \textbf{98.2} & \underline{81.1} \\
+MiMaAlign (ours)   & \textbf{61.0} & \textbf{95.6} & \textbf{61.1} & 98.3 & \textbf{63.9} & \textbf{93.6} & \underline{98.0} & \textbf{81.6} \\
\bottomrule
\end{tabular*}
\vspace{-1em}
\caption{Results on VBench quality score. For each method, the best performance relative to its base model is highlighted in \textbf{bold}.}
\label{tab:vbench_results}
\vspace{-2em}
\end{table*}

\section{Experiments}
\subsection{Experimental Setup}
\textbf{Model Setup.} Our method is built upon the CogVideoX-2B~\cite{Yang25b} text-to-video diffusion model, on top of which we incorporate the proposed MiMaAlign. Except for the newly introduced alignment modules, we keep the original diffusion process and video autoencoder unchanged, and impose auxiliary constraints only in the feature space. This design ensures that our method remains compatible with existing large-scale video generation systems in both architecture and training pipeline.

\textbf{Stage 1: Learning Micro-Level Continuous Dynamics with a Neural ODE.} We use a pre-trained and frozen VideoMAEv2 as the video encoder to extract spatiotemporal features. A Neural ODE is then employed to model the continuous-time dynamics of these features, with Dormand--Prince (Dopri5) adopted as the numerical solver for the ordinary differential equations. This stage is optimized using AdamW with a learning rate of $1\times10^{-4}$ and a batch size of 32.

\textbf{Stage 2: Hierarchical Micro-Macro Alignment.} We fine-tune CogVideoX-2B with hierarchical alignment. The macro-level event teacher is a frozen Qwen2.5-VL 7B model, while the micro-level dynamics teacher is obtained from Stage~1. We insert two lightweight projection heads at different depths of the MM-DiT backbone, applying micro-level dynamics alignment at the 8th layer and macro-level event alignment at the 18th layer. The loss weights are set to $\lambda_{\mathrm{Micro}}=\lambda_{\mathrm{Macro}}=0.5$, and the model is trained with AdamW using a learning rate of $2\times10^{-6}$ and a batch size of 32.

\textbf{Dataset.} We use the same 32K subset of OpenVid~\cite{nan2024openvid} in both training stages. The data split follows the publicly used configuration of VideoREPA to ensure fair comparison under comparable experimental settings.

\textbf{Implementation Details.} All experiments are conducted on four NVIDIA H20 GPUs. More detailed training configurations are provided in the supplementary material.

\subsection{Evaluation}
We conduct a systematic evaluation of the effectiveness of MiMaAlign in improving physical plausibility using two state-of-the-art benchmarks, VideoPhy~\cite{Bansal25a} and VideoPhy2~\cite{Bansal25b}. For general generation quality, we use VBench~\cite{Huang24} for evaluation. We further perform a blind user study to assess the physical plausibility of the generated videos from a human perspective, providing qualitative complementary evidence to the automatic metrics.

To demonstrate its advantages, we compare MiMaAlign against the CogVideoX-2B baseline, other leading open-source text-to-video generation models (HunyuanVideo~\cite{Lab25}, LongCat-Video~\cite{LongCat25}, Cosmos-Predict2.5-2B~\cite{NVIDIA25}, and Wan2.1-14B~\cite{Wan25}), CogVideoX-2B (Finetune) obtained via full-model fine-tuning under the same data and training budget, as well as the recent methods VideoREPA~\cite{Zhang25b} and MoAlign~\cite{Bhowmik25}, both of which are designed to improve physical plausibility. 

\textbf{VideoPhy.} This benchmark focuses on three material-centric interaction categories: solid–solid, solid–liquid, and liquid–liquid. It uses 344 prompts to evaluate Physical Commonsense (PC) and Semantic Adherence (SA), where PC measures physical plausibility and SA measures alignment with the input prompt. We use VideoConPhysics as the automatic scorer and follow the evaluation setup of VideoREPA, testing with augmented prompts. The scoring rule follows WISA~\cite{Wang25} by binarizing the raw score, where scores $\geq 0.5$ are mapped to 1 and otherwise to 0. As shown in Table~\ref{tab:main_results}, applying MiMaAlign to CogVideoX-2B yields substantial improvements in physical plausibility across all categories. Notably, our method achieves the best overall PC score, with a \textbf{relative improvement of 21.0\%} over the CogVideoX-2B baseline. This result highlights the effectiveness of our hierarchical micro-macro alignment framework in capturing complex physical phenomena.

\begin{table}[ht]
\centering
\small
\setlength{\tabcolsep}{2pt}
\renewcommand{\arraystretch}{1.0}
\begin{tabular}{llccc}
\toprule
Comparison &  & MiMaAlign Wins & Baseline Wins & Tie \\
\midrule
\multirow{2}{*}{CogVideoX-2B} 
& SA & \textbf{57.1\%} & 18.9\% & 24.0\% \\
& PC & \textbf{66.9\%} & 20.6\% & 12.6\% \\
\midrule
\multirow{2}{*}{VideoREPA} 
& SA & \textbf{54.3\%} & 20.6\% & 25.1\% \\
& PC & \textbf{59.4\%} & 26.9\% & 13.7\% \\
\midrule
\multirow{2}{*}{MoAlign} 
& SA & \textbf{35.4\%} & 22.3\% & 42.3\% \\
& PC & \textbf{42.3\%} & 30.3\% & 27.4\% \\
\bottomrule
\end{tabular}
\vspace{-1em}
\caption{Blind user study with pairwise Good--Same--Bad (GSB) comparisons. We report the fraction of pairs where MiMaAlign is preferred (Wins), the baseline is preferred, or the two are judged similar (Tie), under Semantic Adherence (SA) and Physical Commonsense (PC) criteria.}
\label{tab:user_study_results}
\vspace{-2em}
\end{table}

\begin{figure}[ht]
    \centering
    \begin{minipage}[b]{0.45\columnwidth}
        \centering
        \small
        \setlength{\tabcolsep}{2pt}
        \renewcommand{\arraystretch}{1.1}
        \begin{tabular}{lcc}
        \toprule
        Methods & SA & PC \\
        \midrule
        Wan2.1-1.3B        & 64.8 & 17.4 \\
        \textbf{+MiMaAlign} & \textbf{65.7} & \textbf{22.1} \\
        \midrule
        CogVideoX-5B       & 67.7 & 31.4 \\
        \textbf{+MiMaAlign} & \textbf{68.3} & \textbf{38.1} \\
        \bottomrule
        \end{tabular}
        \captionsetup{type=table, font=footnotesize}
        \captionof{table}{Ablation Study on Different Diffusion Models. Results of VideoPhy.}
        \label{tab:ablation_models}
    \end{minipage}
    \hfill 
    \begin{minipage}[b]{0.52\columnwidth}
        \centering
        \includegraphics[width=\textwidth]{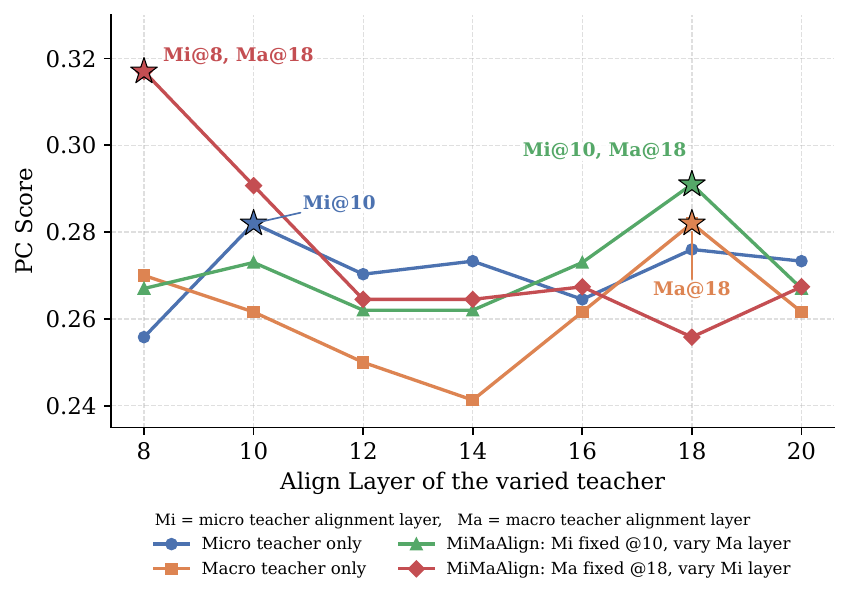}
        \vspace{-2em}
        \captionsetup{type=figure, font=footnotesize}
        \caption{Ablation on teacher usage and alignment depth in terms of PC score on VideoPhy. }
        \label{fig:ablation_pc}
    \end{minipage}
    \vspace{-1.5em}
\end{figure}

\textbf{VideoPhy2.} We further evaluate our model on the more challenging VideoPhy2 benchmark. This benchmark contains 590 detailed prompts describing human--object interactions, covering 200 diverse actions, and uses VideoPhy2-AutoEval as the automatic evaluator. The evaluator reports two metrics: Physical Commonsense (PC) and Semantic Adherence (SA), with each score ranging from 1 to 5. Following the VideoPhy2 protocol, we binarize the raw scores by mapping scores $\geq 4$ to 1 and all others to 0. We use the upsampled prompts provided by the VideoPhy2 benchmark. As shown in Table~\ref{tab:main_results}, our MiMaAlign demonstrates a substantial improvement over the baseline, \textbf{gaining 5.8 points}. We observe that simply adding our training data to CogVideoX-2B does not lead to performance improvements; instead, Physical Commonsense (PC) shows a declining trend. These results indicate that the superior performance of our method does not stem from the training data alone.

\textbf{VBench.} To further verify whether MiMaAlign can improve physical plausibility while maintaining general video generation quality, we report the VBench evaluation results in Table~\ref{tab:vbench_results}. The results show that MiMaAlign maintains highly competitive performance across all dimensions and achieves the highest overall quality score. This demonstrates that our method improves physical plausibility without sacrificing the general video generation quality.

\textbf{User study.} To complement automatic metrics, we conduct a blind user study to evaluate physical plausibility from a human perspective. For each model, we generate 25 videos using augmented prompts sampled from a mixture of VideoPhy and VideoPhy2, and adopt the Good-Same-Bad (GSB)~\cite{Gao2025Seedance} pairwise comparison protocol, which measures relative quality through paired comparisons. As shown in Table~\ref{tab:user_study_results}, participants consistently prefer videos generated by MiMaAlign across all criteria. We provide example videos used in the user study in the supplementary material.

\subsection{Qualitative Comparisons}

Figure~\ref{fig:qualitative} presents qualitative comparisons across models. Compared with CogVideoX-2B, VideoREPA, and MoAlign, MiMaAlign performs best in terms of physical plausibility. Specifically, in the swinging scene, the CogVideoX-2B baseline misses the supporting ropes, which yields an unstable structure. VideoREPA attempts to complete the structure but introduces redundant ropes, and MoAlign produces discontinuous geometry such as broken ropes. In contrast, MiMaAlign generates a complete swing structure with more natural motion. In the spray-painting scene, the paint generated by CogVideoX-2B fails to be properly ejected from the nozzle, violating basic causal and physical processes, while VideoREPA and MoAlign produce anatomically implausible hand distortions when generating humans. MiMaAlign, in contrast, generates videos that better conform to physical commonsense, rendering the spraying action and the material motion process more plausible. For clarity, the figures show shortened versions of the prompts, while the models receive more detailed descriptions during generation. Additional qualitative results are provided in the supplementary material.


\subsection{Ablation Study}
To systematically validate the effectiveness of each core component in the MiMaAlign framework, we conduct a series of ablation studies. All experiments are evaluated on the VideoPhy benchmark, and the results are shown in Figure~\ref{fig:ablation_pc}. We find that aligning micro-level motion features at layer 8 and macro-level physical event features at layer 18 yields the best overall physical performance. In comparison, when only micro-level dynamics alignment is applied, the best PC is achieved at layer 10, while when only macro-level physical event alignment is applied, the best PC is obtained at layer 18. These results provide strong empirical evidence for our central claims: 1) micro-level dynamics and macro-level event knowledge are complementary; and 2) each has its own preferred alignment depth within the generator, and accordingly injecting them into shallower and deeper layers, respectively, leads to better physical performance.

To validate the generalization capability and model-agnostic nature of MiMaAlign, we further apply it to two open-source T2V backbones of different scales and architectures: Wan2.1-1.3B and CogVideoX-5B. As shown in Table~\ref{tab:ablation_models}, MiMaAlign delivers consistent physical performance gains on both backbones, indicating that our method does not depend on a specific model implementation and generalizes well across models.

\section{Conclusion}
In this work, we propose MiMaAlign, a novel hierarchical micro-macro alignment framework designed to systematically improve the physical plausibility of text-to-video diffusion models. We observe that existing methods either rely on discrete motion cues that struggle to capture continuous dynamics, or lack an understanding of high-level physical events. MiMaAlign addresses this issue by introducing two complementary teacher signals that provide supervision at the levels of micro-level motion dynamics and macro-level physical event logic, respectively. The micro-level motion teacher, built on a Neural ODE, captures continuous, fine-grained temporal dynamics, while the macro-level physical event teacher, built on a vision-language model, distills physical commonsense and interaction logic. MiMaAlign aligns these two complementary forms of physical supervision with the shallower and deeper layers of the diffusion backbone, respectively. Extensive experiments show that MiMaAlign achieves significant improvements on multiple physics benchmarks, producing videos with stronger physical plausibility while maintaining general generation quality.

\textbf{Limitations.} Although our method has proven effective, introducing a hierarchical micro-macro alignment framework incurs additional preprocessing and storage overhead, particularly for the macro-level physical event teacher. Nevertheless, this cost is a one-time investment, as the extracted macro-level physical features are stable and reusable for each training sample, making the additional overhead predictable and controllable. Future work could explore more efficient knowledge distillation techniques or adopt smaller, more specialized teacher models to reduce these costs.

\bibliography{aaai2027}


\newpage
\maketitle
\section{Appendix}

\begin{table}[ht]
\centering
\small
\setlength{\tabcolsep}{2pt} 
\renewcommand{\arraystretch}{1.25}
\begin{tabular}{cccccc}
\toprule
Mi@ & Ma@ & Ma Source     & VLM Embedding Layer & SA   & PC   \\
\midrule
8 & 18 & Vision Encoder & --    & \underline{64.2} & 26.7 \\
8 & 18 & LM Decoder     & 0     & 63.4 & 27.3 \\
8 & 18 & LM Decoder     & 1     & 63.1 & 27.6 \\
8 & 18 & LM Decoder     & 15    & 63.4 & \underline{28.2} \\
8 & 18 & LM Decoder     & Final & \textbf{64.8} & \textbf{31.7} \\
\bottomrule
\end{tabular}
\caption{Ablation study comparing VLM visual encoders and language decoders on VideoPhy. The results demonstrate that the hidden states of the language decoder, particularly those from the final layer, provide superior guidance, significantly outperforming representations derived solely from visual features.}
\label{tab:ablation_layer}
\end{table}

\begin{figure}[ht]
    \centering
    \begin{minipage}[b]{0.49\columnwidth}
        \centering
        \includegraphics[width=\textwidth]{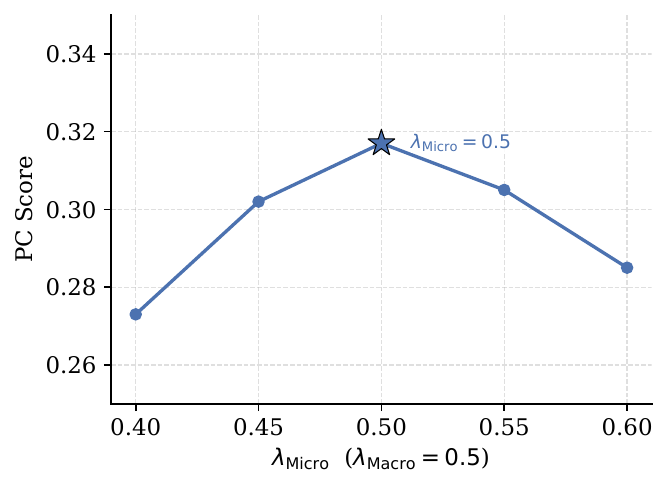}
        \label{fig:ablation_pc_lambda_micro}
    \end{minipage}
    \hfill
    \begin{minipage}[b]{0.49\columnwidth}
        \centering
        \includegraphics[width=\textwidth]{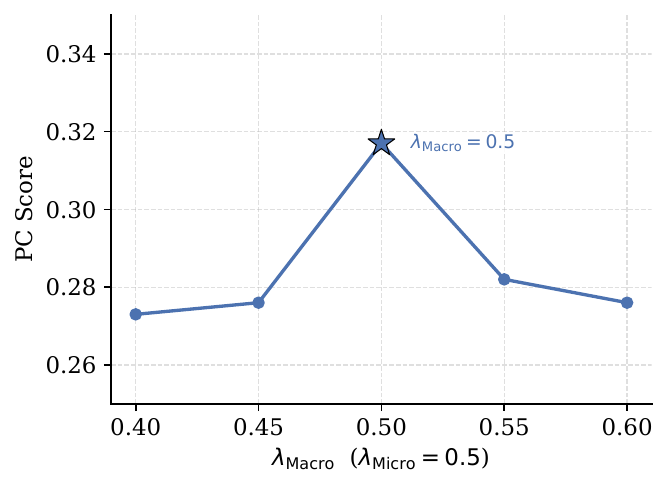}
        \label{fig:ablation_pc_lambda}
    \end{minipage}
    \vspace{-1em}
    \caption{Ablation on the alignment weight $\lambda$ in terms of the PC score on VideoPhy. Left: fixing $\lambda_{\text{Macro}}=0.5$ and varying $\lambda_{\text{Micro}}$; right: fixing $\lambda_{\text{Micro}}=0.5$ and varying $\lambda_{\text{Macro}}$.}
    \label{fig:ablation_pc_lambda_all}
    \vspace{-1.5em}
\end{figure}

\subsection{Additional ablation study}


To further analyze where the gains brought by the macro-level physical event teacher come from, we perform a finer-grained ablation on the source of its internal representations (Table~\ref{tab:ablation_layer}). With the micro-level alignment layer fixed at layer 8 (Mi@$=8$) and the macro-level alignment layer fixed at layer 18 (Ma@$=18$), using only the output of the VLM vision encoder as macro-level supervision (Vision Encoder) yields a considerable improvement, but it is consistently weaker than aligning to the hidden states of the VLM language decoder (LM Decoder). As the selected decoder layer becomes deeper, the PC score generally increases, and the final-layer representation achieves the best results, which is consistent with the design in the main paper that adopts the visual tokens of the VLM's final-layer hidden states as the macro-level alignment target. This pattern indicates that, compared with purely visual representations, higher-level decoder states, refined through instruction-guided multimodal reasoning, better encode object interactions, causal dependencies, and event-level physical commonsense. They therefore provide a stronger macro-level event supervision signal for the diffusion model, further promoting physical plausibility while improving semantic consistency.

\textbf{Effect of the alignment weight $\lambda$.} We further examine the effect of the alignment weights on performance by separately varying the weights of the micro- and macro-level losses, $\lambda_{\text{Micro}}$ and $\lambda_{\text{Macro}}$. As shown in Figure~\ref{fig:ablation_pc_lambda_all}, $\lambda_{\text{Micro}}=\lambda_{\text{Macro}}=0.5$ achieves the best balance between the original diffusion loss and the micro- and macro-level losses.

\subsection{Physion evaluation setting}
In this paper, in order to evaluate and validate the effectiveness of the features extracted by our Neural-ODE-based micro-level motion teacher for physical understanding, we use the Physion benchmark~\cite{Bear2021Physion}. The Physion benchmark provides realistic simulations of diverse physical scenes, in which objects are manipulated under different configurations to assess the ability of a model to perform various types of physical reasoning such as stability, rolling, and linking. We adopt its latest version, Physion v1.5~\cite{Bear2021Physion}. Since Physion contains diverse physical phenomena, complex object dynamics, and realistic 3D simulations, it serves as a highly challenging benchmark.

Our main goal is to show that the motion features learned under Neural ODE supervision have a stronger ability to understand the physical world than those learned with conventional optical flow supervision. To this end, we design a comparative experiment. Specifically, we first use a pretrained and frozen VideoMAEv2~\cite{Wang23} as the base video encoder, and then train a motion projection head under two different supervision schemes to disentangle dedicated motion representations from the mixed features of VideoMAEv2. The first scheme is our proposed Neural ODE based method, which performs self-supervised learning by predicting the continuous-time evolution of features. The second scheme follows the approach in MoAlign~\cite{Bhowmik25} and uses optical flow prediction as the supervision signal. After feature extraction, we evaluate these motion features on the Object Contact Prediction (OCP) task in Physion. The OCP task requires the model to predict, given an initial context video, whether two objects will make contact in the future, which demands a strong implicit understanding of physical dynamics.

We first use the two motion feature extractors trained with the above methods to extract motion features from the Physion test videos. From the twenty-four temporal dimensions available in each type of motion feature, we uniformly sample six temporal dimensions and use all spatial tokens within these temporal slices. The extracted motion features are then used to train a logistic regression classifier to perform the OCP task. We evaluate on four Physion subsets, link, roll, contain, and dominoes, and we report the prediction accuracy in Table~1 of the main text.



\subsection{Stage 1 Training Details}
\label{subsec:stage1_details}
To learn a micro-level motion representation that captures fine-grained, continuous-time dynamics, we build a self-supervised learning framework based on Neural Ordinary Differential Equations (Neural ODEs), corresponding to Stage 1 in the main paper. The input video is first resized to a resolution of $160 \times 240$ and truncated to $49$ frames. Then, we use a pretrained and frozen VideoMAEv2 encoder $\mathcal{V}$ to extract spatiotemporal token-level features with dimension $D_v = 768$ from the input video, and we reshape them into a feature sequence of shape $[B, 768, 24, H', W']$. After that, we compress the features with a lightweight temporal projection head $\Phi_\psi$. Specifically, the module consists of two causal Conv1D layers along the temporal axis, with kernel size $3$ and padding only on the left to preserve temporal causality. The first layer compresses the channel dimension from $768$ to $256$ and applies a GELU activation; the second layer further compresses the channel dimension to $D_m = 128$, thereby mapping the encoder output into a $128$-dimensional motion subspace and producing features $f_{\text{micro}}$ of shape $[B, 128, 24, H', W']$.

Based on this, we use a Neural ODE module to characterize the continuous evolution of the motion features over time within the motion subspace. Specifically, we take the temporal axis as the integration axis and treat the feature vectors in the motion subspace as states evolving over time, whose variation is described by a feedforward network parameterizing the dynamics equation $\mathrm{d}z/\mathrm{d}t = \mathcal{F}_\theta(z, t)$, implemented as an MLP with two linear layers and a GELU activation in between. We use the Dormand--Prince (Dopri5) numerical solver $\mathrm{ODESolve}(\cdot)$ to integrate the resulting differential equation. During training, we adopt a self-supervised sliding-window prediction strategy: at each time step $t$, we take the current motion features $f_{\text{micro}}^{(t)}$ as the initial state and solve the Neural ODE to predict the features $\hat{f}_{\text{micro}}^{(t+\tau)}$ at the next $N = 4$ temporal offsets $\tau = 1,\dots,N$, and match them with the ground-truth motion features extracted from the same video at the corresponding future times, forming the self-supervised feature prediction objective $\mathcal{L}_{\text{phys}}$. Minimizing this objective forces the projection head and the dynamics network to jointly learn the underlying continuous dynamics in videos, yielding the motion representation $F_{\text{Micro}}$ used for micro-level alignment in Stage 2. This stage is trained with the AdamW optimizer using a learning rate of $1 \times 10^{-4}$ and a batch size of $32$, and it runs for $1$ epoch. Training is performed on four NVIDIA H20 GPUs, each with $96$GB of memory. In Stage 2, both the VideoMAEv2 encoder and this motion projection head are frozen and serve only as the micro-level dynamics teacher to provide supervision targets.

\subsection{Stage 2 Training Details}
\label{subsec:stage2_details}
\begin{figure*}[t]
    \centering
    \includegraphics[width=1\textwidth]{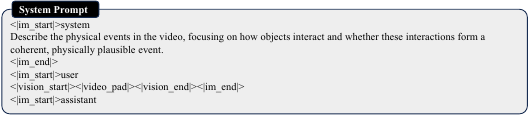} 
    \caption{System prompt used to guide the macro-level physical event teacher (Qwen2.5-VL) during alignment.}
    \label{fig:System_Prompt}
\end{figure*}

In the second stage, we fully fine-tune the pretrained CogVideoX-2B model and inject the complementary knowledge of the micro-level dynamics teacher and the macro-level physical event teacher into different depths of the diffusion backbone through hierarchical alignment. The micro-level dynamics teacher comes from Stage 1, while the macro-level physical event teacher is a pretrained, frozen, and instruction-prompted Qwen2.5-VL 7B model; we use a system prompt to guide it to focus on object interactions, causal dependencies, and event-level physical commonsense, and the complete prompt is provided in Figure~\ref{fig:System_Prompt}. Following the design in the main paper, we attach two independent lightweight MLP projection heads to the MM-DiT backbone (corresponding to $P_i$ and $P_j$ in the main paper). The micro-level alignment head is connected to the $8$th MM-DiT layer and maps the $1920$-dimensional features into the $D_m = 128$-dimensional motion subspace; the macro-level alignment head is connected to the $18$th MM-DiT layer and maps the features into a $D_s = 3584$-dimensional event representation space aligned with the hidden states of Qwen2.5-VL.

Since the student and teacher features differ in spatiotemporal resolution, we apply a unified resolution alignment procedure to the student features before computing the alignment losses. Along the temporal dimension, we upsample the student features with trilinear interpolation so that they match the frame rate of the teacher features. Along the spatial dimensions, we downsample the student features with a $3 \times 3$ convolution of stride $3$ to match the spatial resolution of the teacher features.

During training, we freeze the Qwen2.5-VL 7B macro-level physical event teacher, the VideoMAEv2 encoder, and the motion projection head $\Phi_\psi$ learned in Stage 1, and optimize only the CogVideoX backbone transformer and the two alignment projection heads. Both alignment projection heads are implemented as single-layer MLPs with SiLU activation. The loss weights are set to $\lambda_{\mathrm{Micro}} = \lambda_{\mathrm{Macro}} = 0.5$, consistent with the main paper. In the TRD loss, we use a margin parameter $m = 0.1$ and clamp loss values below this margin to $0$. The fine-tuning stage uses the AdamW optimizer with a learning rate of $2 \times 10^{-6}$ and a batch size of $32$. Training is conducted on four NVIDIA H20 GPUs.

\textbf{Transfer to other backbone models.} To verify the generalization ability and model-agnostic nature of MiMaAlign, we further transfer it to two open-source T2V backbones with different scales and architectures, namely CogVideoX-5B and Wan2.1-1.3B. For both models, we use LoRA for efficient fine-tuning over $1000$ steps on $32$K OpenVid videos. Unless otherwise specified, the teacher settings, alignment losses, data split, and most training hyperparameters remain the same as those used for CogVideoX-2B.

By default, micro-level dynamics alignment is applied at the $8$th layer and macro-level event alignment at the $18$th layer. For LoRA fine-tuning on CogVideoX-5B, we set the learning rate to $1 \times 10^{-4}$, the LoRA rank to $128$, and alpha to $64$. For Wan2.1-1.3B, we set the learning rate to $2 \times 10^{-6}$, the LoRA rank to $4$, and alpha to $4$. All experiments are conducted on four NVIDIA H20 GPUs with a total batch size of $32$.



\subsection{User Study}
To complement the automatic metrics from a human perspective, we conduct a blind user study. Participants watch $25$ videos generated by four models from augmented prompts sampled from a mixed set of VideoPhy~\cite{Bansal25a} and VideoPhy2~\cite{Bansal25b}. We conduct pairwise comparisons between MiMaAlign and CogVideoX-2B~\cite{Yang25b}, VideoREPA~\cite{Zhang25b}, and MoAlign~\cite{Bhowmik25}, where all compared methods are built on the CogVideoX-2B backbone. We adopt the Good--Same--Bad (GSB)~\cite{Gao2025Seedance} protocol: for each video pair, participants, without knowing which model each video comes from, judge which video is better according to the following two criteria, or declare the two similar (Tie).
\begin{itemize}
    \item \textbf{Semantic Adherence (SA)}: Which video better aligns with the text prompt?
    \item \textbf{Physical Commonsense (PC)}: Which video appears more physically realistic (e.g., regarding causality, object motion, fluid dynamics)?
\end{itemize}

\subsection{More qualitative results}
\begin{figure*}[t]
    \centering
    \includegraphics[width=0.9\textwidth]{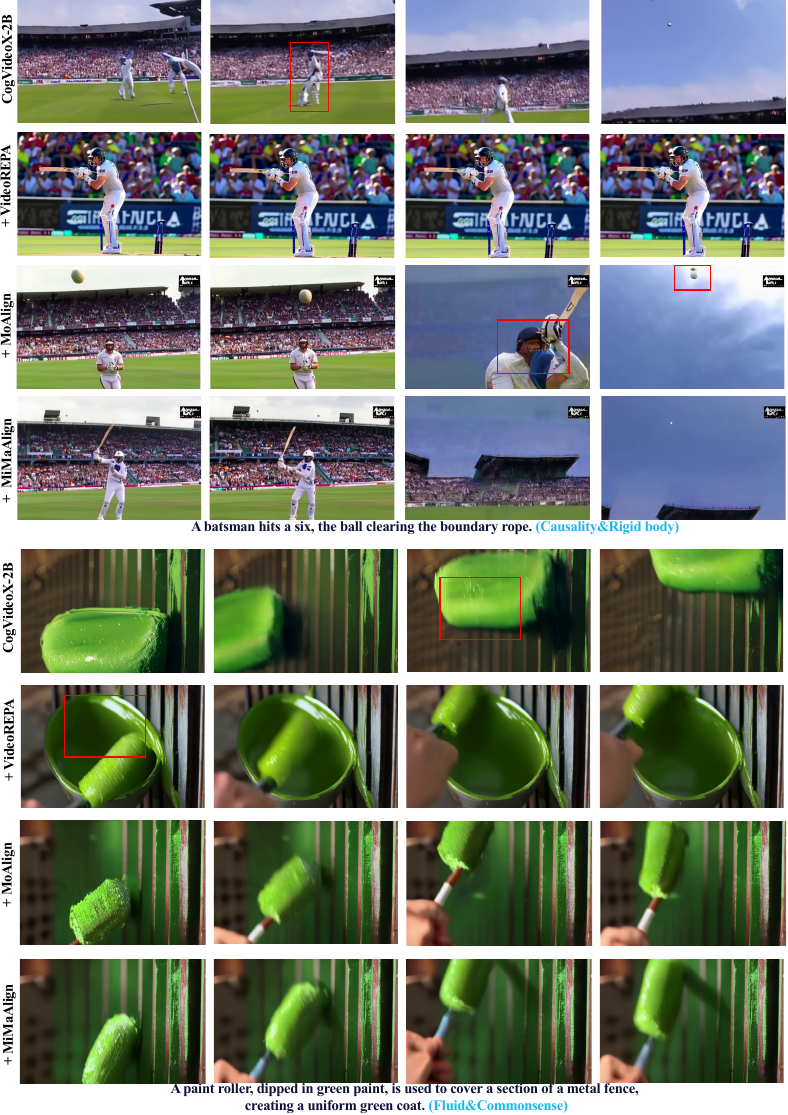} 
    \caption{Qualitative comparison of MiMaAlign against the CogVideoX-2B baseline, VideoREPA, and MoAlign, all built on the CogVideoX-2B backbone.}
    \label{fig:cog_videorepa_moalign_mima}
\end{figure*}

\begin{figure*}[t]
    \centering
    \includegraphics[width=0.9\textwidth]{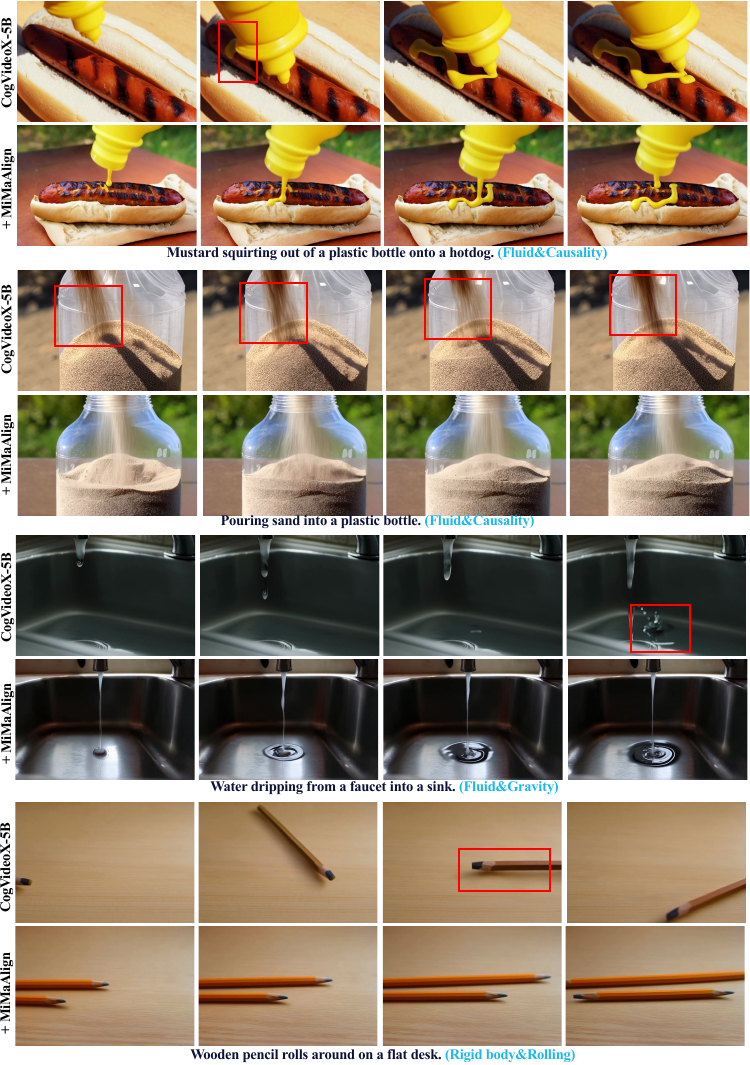} 
    \caption{Qualitative comparison. The first row shows the outputs of CogVideoX-5B, and the second row presents the results of our MiMaAlign.}
    \label{fig:cog_5b_mima}
\end{figure*}

\begin{figure*}[t]
    \centering
    \includegraphics[width=0.9\textwidth]{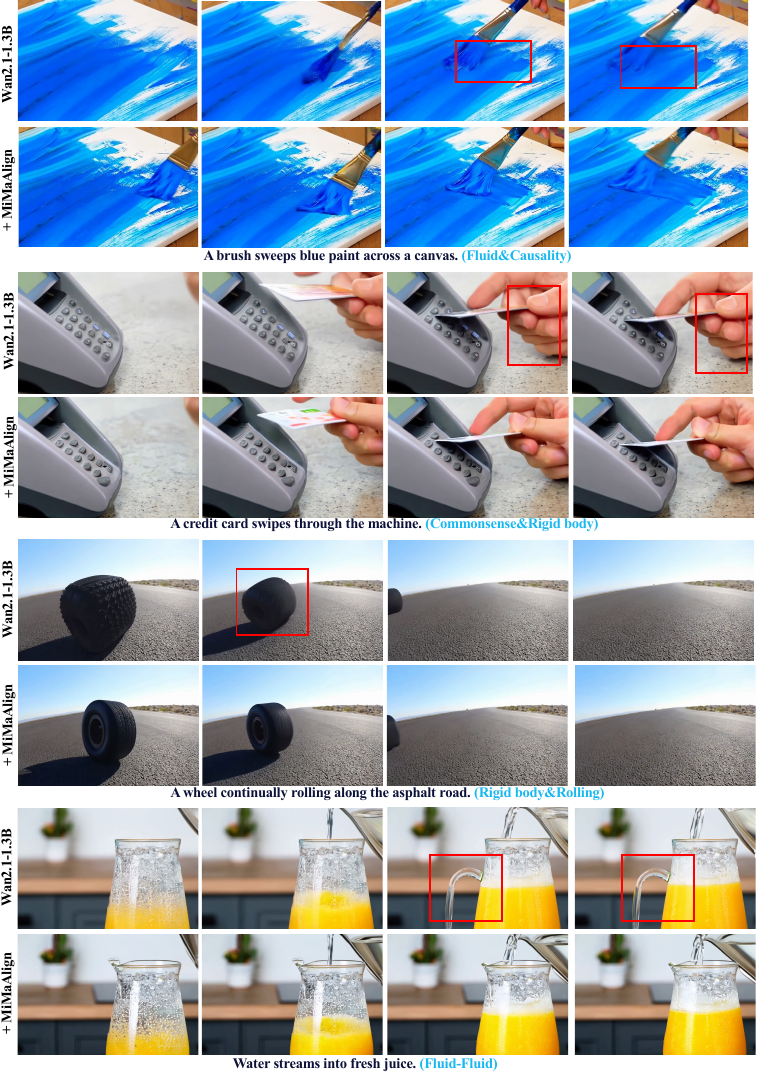} 
    \caption{Qualitative comparison. The first row shows the outputs of Wan2.1-1.3B, and the second row presents the results of our MiMaAlign.}
    \label{fig:wan_mimr_480_720}
\end{figure*}

\begin{figure*}[t]
    \centering
    \includegraphics[width=0.9\textwidth]{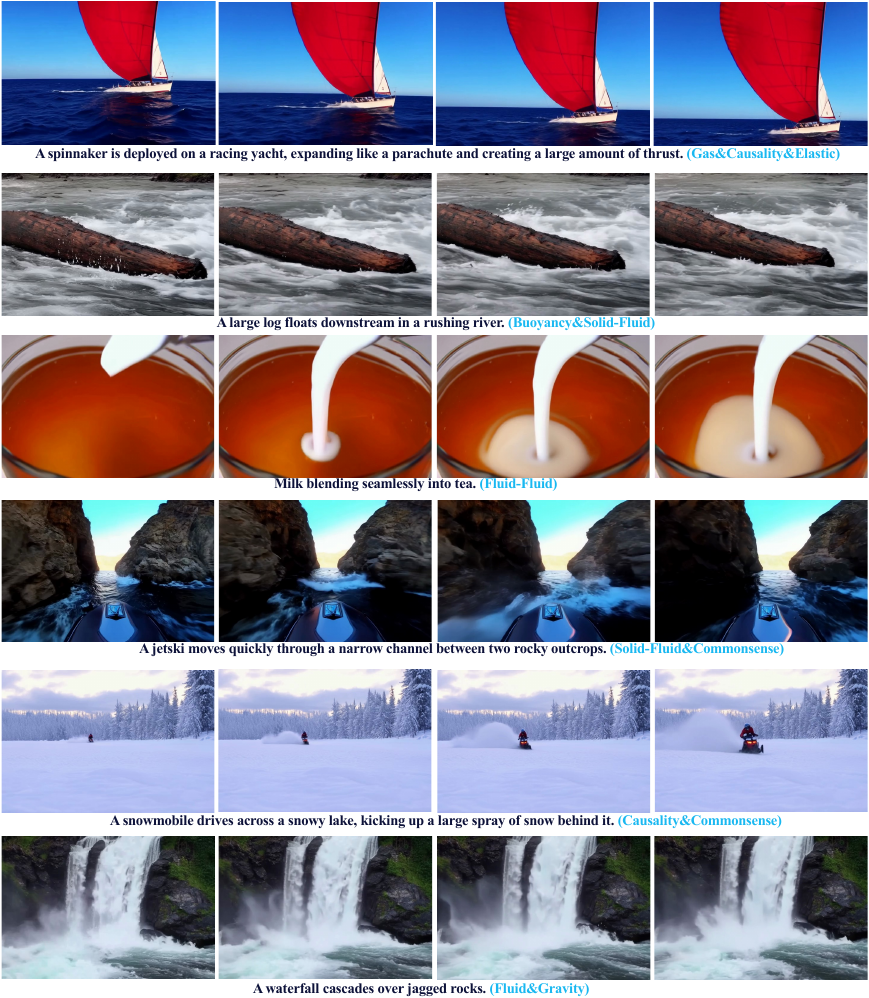} 
    \caption{Qualitative generation results of MiMaAlign based on CogVideoX-2B.}
    \label{fig:single_mima_cog_2B}
\end{figure*}

\begin{figure*}[t]
    \centering
    \includegraphics[width=0.9\textwidth]{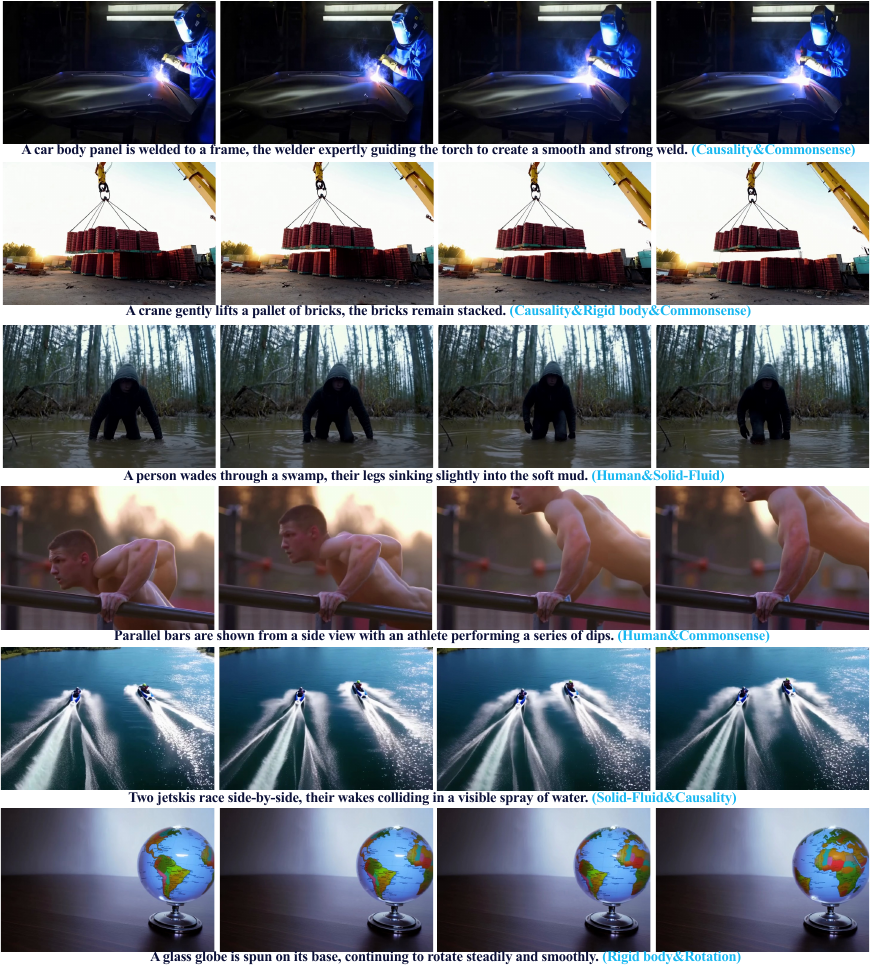} 
    \caption{Qualitative generation results of MiMaAlign based on CogVideoX-5B.}
    \label{fig:single_mima_cog_5B}
\end{figure*}

\begin{figure*}[t]
    \centering
    \includegraphics[width=0.9\textwidth]{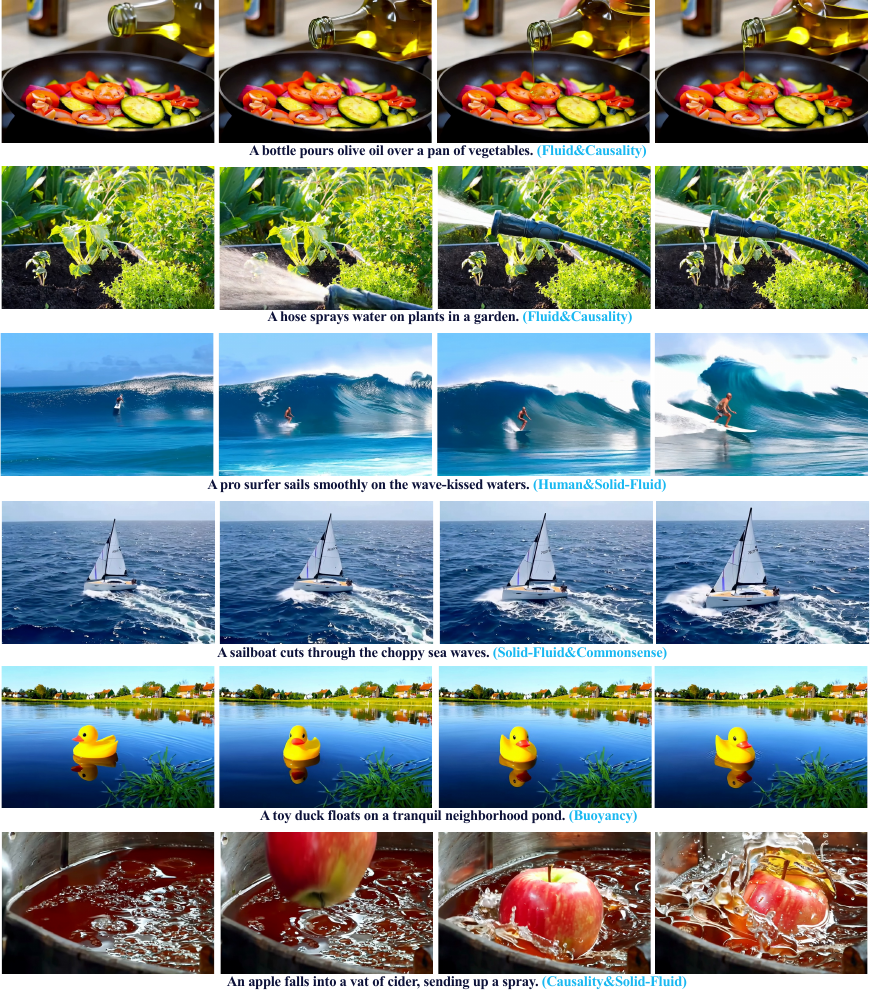} 
    \caption{Qualitative generation results of MiMaAlign based on Wan2.1-1.3B.}
    \label{fig:single_mima_wan1}
\end{figure*}

In this section, we present more video generation results of MiMaAlign. Figure~\ref{fig:cog_videorepa_moalign_mima} directly compares MiMaAlign with the CogVideoX-2B baseline, VideoREPA, and MoAlign on the CogVideoX-2B backbone, while Figures~\ref{fig:cog_5b_mima} and~\ref{fig:wan_mimr_480_720} show the improvements of MiMaAlign over the CogVideoX-5B and Wan2.1-1.3B baselines, respectively. These comparisons further demonstrate that our method enhances the physical commonsense of generated videos across backbones of different scales and architectures. In addition, to further showcase its capability, we present in Figures~\ref{fig:single_mima_cog_2B}, \ref{fig:single_mima_cog_5B}, and~\ref{fig:single_mima_wan1} more videos generated by MiMaAlign based on CogVideoX-2B, CogVideoX-5B, and Wan2.1-1.3B, respectively, which exhibit strong physical plausibility. Red rectangles in the figures mark phenomena that violate physical commonsense for easier identification.

\subsection{Text prompts used for generation}
\begin{figure*}[t]
    \centering
    \includegraphics[width=0.9\textwidth]{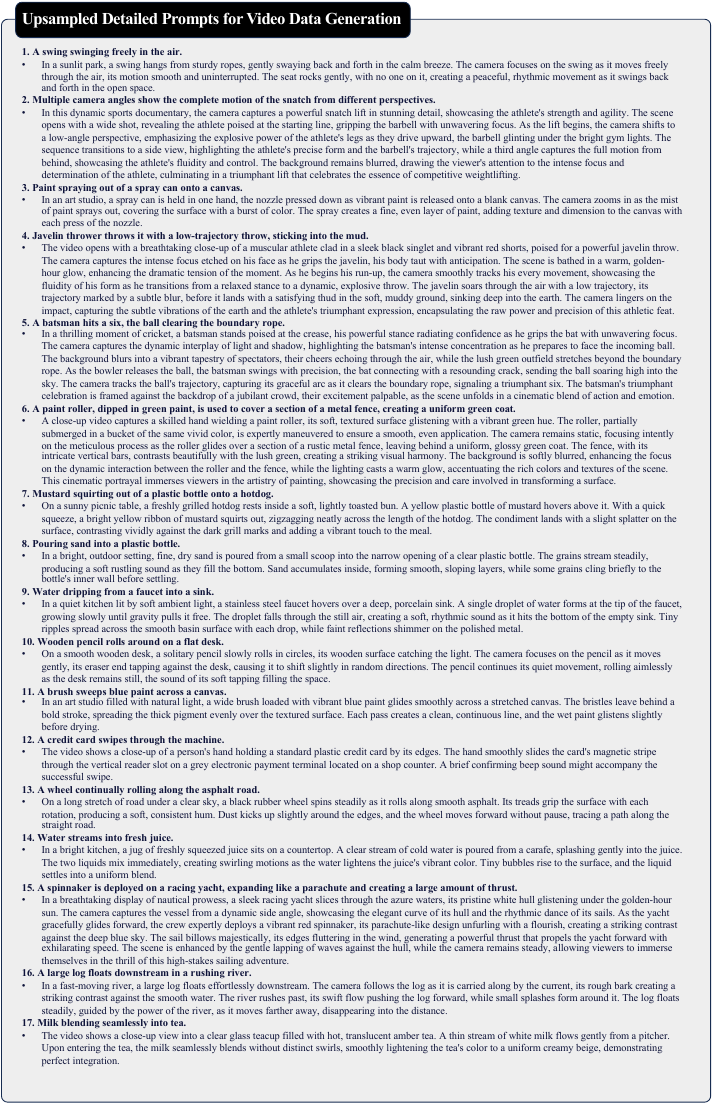} 
    \caption{Upsampled Detailed Prompts for Video Data Generation.}
    \label{fig:long_prompt1}
\end{figure*}

\begin{figure*}[t]
    \centering
    \includegraphics[width=0.9\textwidth]{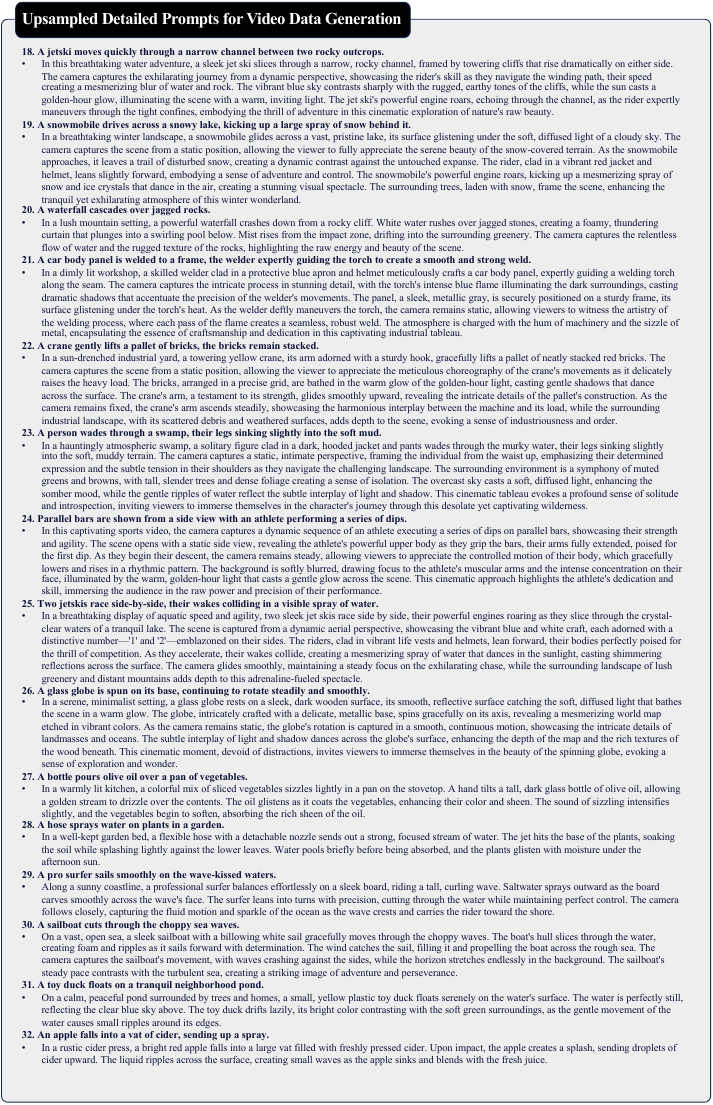} 
    \caption{Upsampled Detailed Prompts for Video Data Generation.}
    \label{fig:long_prompt}
\end{figure*}

In this section, Figures~\ref{fig:long_prompt1} and~\ref{fig:long_prompt} provide the complete text prompts used to generate the videos for the qualitative comparisons.

\end{document}